\documentclass[twocolumn]{autart}  
\usepackage{cite}
\usepackage{amsmath,amssymb,amsfonts}
\usepackage{algorithm}
\usepackage{algpseudocode}
\usepackage{graphicx}

\usepackage{textcomp}
\usepackage{xcolor}
\usepackage{soul,dsfont} 
\usepackage{tikz} 
\usepackage{color}
\usepackage{algorithm} 
\usepackage{array}
\usepackage{eqparbox}
\usepackage{url}
\usepackage{stfloats}
\usepackage{graphics}
\usepackage[mathscr]{euscript}
\graphicspath{{figures/}{./}}

\newtheorem{theorem}{Theorem}
\newtheorem{assumption}{Assumption}

\begin{document}

\begin{frontmatter}

\title{Higher-Order Interaction Reconstruction via Integral Physics-Informed Neural Networks\thanksref{footnoteinfo}}

\thanks[footnoteinfo]{This paper was not presented at any IFAC meeting. Xin Mao and Zhen Xu contributed equally to this work. Can Chen is supported by the National Science Foundation under Grant CMMI-2611908.  Corresponding authors: Tianlong Chen; Can Chen.}

\author[1]{Xin Mao},
\author[2]{Zhen Xu},
\author[3]{Shiyi Yang},
\author[2]{Sukwon Yun},
\author[4]{Yunmei Liu},
\author[5]{Ren Wang},
\author[2]{Tianlong Chen}\ead{tianlong@cs.unc.edu},
\author[1,3,6]{Can Chen}\ead{canc@unc.edu}

\address[1]{School of Data and Information Sciences, University of North Carolina at Chapel Hill, Chapel Hill, NC 27599, USA} 
\address[2]{Department of Computer Science,  University of North Carolina at Chapel Hill, Chapel Hill, NC 27599, USA}
\address[3]{Department of Biostatistics, University of North Carolina at Chapel Hill, Chapel Hill, NC 27599, USA}
\address[4]{Department of Industrial \& Systems Engineering,  University of Louisville, Louisville, KY 40292, USA }
\address[5]{Department of Electrical and Computer Engineering, Illinois Institute of Technology, Chicago, IL 60616, USA}
\address[6]{Department of Mathematics, University of North Carolina at Chapel Hill, Chapel Hill, NC 27599, USA}


\begin{abstract}
Reconstructing higher-order interactions  is a fundamental challenge in complex networked systems, with  applications in ecological, social, and biological networks. This inverse problem is inherently difficult due to the combinatorial complexity of higher-order structures, the strong nonlinearity of the underlying dynamics, and the prevalence of noisy and sparse observations. In this article, we propose a novel integral physics-informed neural network (IntegralPINN) framework for higher-order interaction reconstruction. Instead of { penalizing derivative-based physics residuals,} we adopt an integral formulation over temporal intervals, which naturally suppresses high-frequency noise and mitigates derivative-induced instability. We further introduce a continuous parameterization of  interaction tensors, enabling tractable identification of higher-order  structures via gradient-based optimization.  We provide theoretical analysis establishing a noise-robust error bound that is independent of derivative amplification { when the integration window satisfies an excitation condition.} 
Extensive experiments on ecological dynamics, social contagion, and neuronal synchronization systems demonstrate that IntegralPINN achieves accurate and robust reconstruction under noisy and sparse observations, outperforming  state-of-the-art approaches.
	\end{abstract}
    \end{frontmatter}

\section{Introduction}\label{sec:intro}
Higher-order interactions play a fundamental role in the emergence of collective behavior in complex  systems, spanning ecological, biological, social, and mechanical networks \cite{battiston2021physics,billick1994higher,gao2023dynamics,mayfield2017higher,malizia2024reconstructing,chen2024stability,kleinhesselink2022detecting,wang2024epidemic,chen2021controllability}. In such systems, models based solely on pairwise interactions often fail to capture the inherently multi-body effects. For example, ecological competition may involve multiple species simultaneously, information adoption can depend on group reinforcement, and neuronal dynamics may be shaped by multi-node synchronization patterns \cite{mickalide2019higher,yu2011higher,aladwani2019addition,mao2026model,levine2017beyond,majhi2022dynamics}.
Motivated by these observations, there has been growing interest in modeling networked dynamical systems with higher-order interactions, where the evolution of each node depends on joint interactions among multiple nodes \cite{millan2025topology,shi2024hypergraph,niedostatek2025mining,ferraz2024contagion}. A central challenge is to reconstruct the underlying higher-order interaction structure from observed time-series data, which is essential for downstream tasks such as understanding, prediction, and control of complex systems. 

This inverse problem is fundamentally difficult due to both structural and dynamical complexities. The number of possible higher-order interactions grows combinatorially with the system size. Moreover, the interaction structure is tightly coupled with nonlinear system dynamics, while noisy and sparse observations further complicate reliable reconstruction. A variety of approaches have been proposed for reconstructing higher-order interactions from data, including correlation-based methods \cite{kettenring1971canonical}, regression and kernel-based techniques \cite{donati2026kernel}, statistical inference and validation frameworks \cite{elmes2022fast,lizotte2023hypergraph,young2021hypergraph,kritschgau2024community,musciotto2021detecting}, sparsity-promoting optimization methods \cite{delabays2025hypergraph,zang2024stepwise}, and dynamical system-based approaches that exploit time-series observations \cite{mao2025tensor,tabar2024revealing,mao2026data}. { More broadly, sparse nonlinear identification methods, such as SINDy \cite{brunton2016discovering} and its integral and weak-form variants \cite{schaeffer2017sparse,messenger2021weak}, can also be applied to this problem by identifying active terms from a prescribed library of candidate functions.} While the above methods can be effective under certain conditions, they typically rely on statistical dependencies or simplified modeling assumptions, and may struggle to distinguish genuine higher-order interactions from indirect effects. { Methods that exploit temporal dynamics  face additional challenges, as numerical differentiation can amplify measurement noise, while derivative-free formulations, including integral and weak-form SINDy, remain susceptible to errors in nonlinear features constructed from noisy observations.}

Recently, a neural network approach named hypergraph physics-informed neural network (HyperPINN) has been proposed to jointly learn system trajectories and hypergraph structures by combining data-fitting, physics-informed losses, and sparsity regularization, and demonstrates state-of-the-art performance for higher-order interaction reconstruction \cite{mao2025identification}. However, its effectiveness degrades as the interaction order increases, due to the combinatorial growth of higher-order terms, which leads to a rapidly expanding parameter space and increased optimization difficulty.
More fundamentally, standard PINN formulations { encourage physical consistency by penalizing differential residuals}, where the time derivatives of the system state are matched against model predictions \cite{raissi2019physics,huang2022applications,antonelo2024physics,norambuena2024physics,nicodemus2022physics}. Hence, computing such derivatives { from a surrogate fitted to} noisy and discretely sampled observations is inherently unstable and can lead to significant noise amplification, especially in the presence of nonlinear, higher-order  interactions. Moreover, for non-smooth or event-driven processes, such as binary state transitions or switching dynamics the underlying derivatives may be not well-defined or numerically unstable, rendering such approaches unreliable. { To alleviate these limitations, weak-form PINNs integrate physics residuals against test functions \cite{kharazmi2019variational,kharazmi2021hp,zang2020weak}, thereby reducing  derivative-related noise sensitivity and accommodating certain non-smooth solutions. Nevertheless, these formulations require appropriate choices of test functions and numerical quadrature, and have primarily been developed for solving partial differential equations with prescribed structures rather than reconstructing unknown higher-order interactions from a combinatorially large candidate space.}

Here, we propose a novel integral physics-informed neural network (IntegralPINN) framework for higher-order interaction reconstruction from discrete and noisy observations. Instead of { penalizing differential residuals}, we { promote} physical consistency via an integral formulation over temporal intervals, replacing derivative matching with state increments and integrated vector fields. We represent the interaction structure using learnable interaction tensors, which are jointly optimized with a continuous neural surrogate of the system trajectory, enabling simultaneous dynamics estimation and structure identification. Beyond the algorithmic design, we provide a theoretical analysis showing that derivative-based PINN formulations incur an additional error term due to high-frequency artifacts introduced by automatic differentiation. In contrast, the proposed integral formulation eliminates this amplification and yields parameter estimation errors that scale { with the approximation accuracy of the surrogate model when the integration window satisfies an excitation condition}. Furthermore, we interpret the integral operator as a spectral preconditioner that suppresses high-frequency components, providing a frequency-domain explanation for the improved stability.  IntegralPINN is validated on small- and medium-scale ecological dynamics, social contagion, and neural synchronization systems, demonstrating strong robustness to noise and the ability to handle both continuous dynamics and non-smooth processes, including binary observations and event-driven behaviors.

{ Unlike purely statistical methods, IntegralPINN exploits the known governing dynamics to help distinguish genuine higher-order interactions from indirect effects. Compared with  regression-based approaches such as SINDy or integral SINDy, it jointly estimates the state trajectory and the interaction structure, allowing the governing dynamics to regularize trajectory estimation and mitigate the effects of observation noise.
Within physics-informed learning, its integral residuals mitigate the differentiation-induced amplification of surrogate errors in HyperPINN, while avoiding the need for a prescribed test-function basis in weak-form PINNs and directly targeting unknown interaction structures.} The remainder of this article is organized as follows. Section~\ref{sec:problem} presents the problem formulation of higher-order interaction reconstruction. Section~\ref{sec:methodology} introduces the proposed IntegralPINN framework. Section~\ref{sec:theory} provides theoretical analysis of error bounds. Section~\ref{sec:example} presents numerical experiments, and  Section~\ref{sec:conclusion} concludes the article with a discussion of future directions.

\section{Problem Formulation}
\label{sec:problem}
\begin{figure}[t]
    \centering
\includegraphics[width=0.5\linewidth]{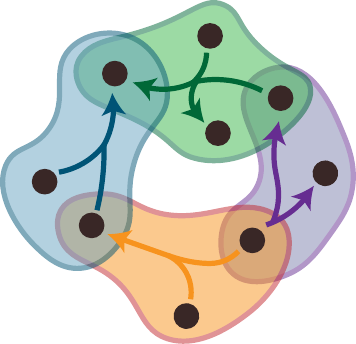}
    \caption{An example of higher-order interactions in a network.}
    \label{fig:hypergraph}
\end{figure}

A network with higher-order interactions is described as a node set $\mathcal{V} = \{v_1, v_2, \dots, v_n\}$ and a collection of higher-order interactions $\mathcal{E}$, where each element $e \in \mathcal{E}$ satisfies $e \subseteq \mathcal{V}$. To capture directional dependencies, each interaction $e$ is specified as an ordered pair $e = (e^h, e^t)$, where $e^h, e^t \subseteq \mathcal{V}$ denote the head and tail sets, respectively, and satisfy $e^h \cap e^t = \emptyset$. The size of each interaction is upper bounded by a prescribed maximum order $K$. An illustrative example is provided in Fig.~\ref{fig:hypergraph}.
For each interaction order $k = 2,3, \dots, K$, we define an interaction tensor $\mathscr{B}^{(k)} \in \mathbb{R}^{n \times n\times \stackrel{k}{\cdots} \times n}$ as
\begin{equation*}
\mathscr{B}^{(k)}_{j_1 j_2 \cdots j_k}
=
\begin{cases}
c, & \text{if } (\{v_{j_1}\}, \{v_{j_2},v_{j_3}, \dots, v_{j_k}\}) \in \mathcal{E}, \\
0, & \text{otherwise},
\end{cases}
\end{equation*}
where $c$ denotes the interaction weight. Although the above definition is written in a single-head form, interactions with multiple head nodes can be represented by assigning nonzero entries corresponding to each head node separately. Consequently, nonzero entries of $\mathscr{B}^{(k)}$ indicate the presence of directed higher-order interactions and quantify their strengths.

Let $x_i \in \mathbb{R}$ denote the state of node $v_i$, and define $\textbf{x} = [x_1 \ x_2 \ \cdots \ x_n]^\top$. The dynamics of the network with higher-order interactions is described as
\begin{align*}
\scalebox{0.95}{
$\displaystyle\dot{x}_i = f(x_i)
+ \sum_{k=2}^{K} \sum_{j_2=1}^n\cdots \sum_{j_k=1}^n
\mathscr{B}^{(k)}_{i j_2\cdots j_{k}}f^{(k)}(x_i, x_{j_2}, \dots, x_{j_{k}}),$}
\end{align*}
where $f(\cdot)$ represents the local node dynamics, and $f^{(k)}(\cdot)$ denotes the nonlinear mapping associated with $k$th-order interactions.  
In practice, the system states are not directly observed. Instead, we are given discrete and noisy measurements
\begin{equation*}
\tilde{\textbf{x}}(t_l) = \textbf{x}(t_l) + \textbf{w}(t_l),\quad l = 1, 2,\dots, M,
\end{equation*}
where $\textbf{w}(t_l)$ denotes the measurement noise. The collected data matrix is denoted as
\begin{equation*}
\tilde{\textbf{X}} =
\begin{bmatrix}
\tilde{\textbf{x}}(t_1) & \tilde{\textbf{x}}(t_2) & \cdots & \tilde{\textbf{x}}(t_M)
\end{bmatrix}.
\end{equation*}
As the interaction order $k$ increases, the number of possible higher-order interactions grows combinatorially with the number of nodes $n$, leading to a high-dimensional and potentially sparse structure identification problem.

We aim to reconstruct the underlying higher-order interaction structure from the observed time-series data $\tilde{\textbf{X}}$. To render the problem tractable, we { impose the following assumption.}

\begin{assumption}\label{ass:known}
{ The local node dynamics $f(\cdot)$ and the functional forms of the higher-order interaction mappings $f^{(k)}(\cdot)$, $k=2,3,\ldots,K$, are known.} 
\end{assumption}

{ This assumption is appropriate for systems where the governing interaction mechanisms are available from established physical or mechanistic models, while the underlying interaction structure remains unknown. Examples include ecological population models, epidemic spreading processes, and coupled oscillator networks.}
Similar assumptions on known interaction functions are adopted in model-based higher-order interaction reconstruction methods \cite{mao2025identification,mao2025tensor}.
The main challenge lies in recovering the high-dimensional and sparse interaction tensors $\mathscr{B}^{(k)}$ from noisy observations while ensuring consistency with the underlying nonlinear dynamics, which results in a highly nontrivial inverse problem. { Although the dynamics is linear in the entries of the interaction tensors under Assumption~\ref{ass:known}, the regression features $f^{(k)}(\cdot)$ depend on the underlying state trajectory, which is not directly observed, so that evaluating the features from the observations introduces errors that propagate through the nonlinear higher-order interaction terms.} This motivates the development of a robust framework that jointly learns a continuous surrogate of the state trajectory and the interaction structure, avoids explicit differentiation, and enables stable identification from such observations.

\section{Integral Physics-Informed Neural
Networks}\label{sec:methodology}
In this section, we present the proposed integral physics-informed neural network (IntegralPINN) framework for reconstructing higher-order interaction structures from discrete and noisy observations. The key idea is to replace the derivative-based physics { residuals} used in conventional PINNs with an integral formulation, thereby { encouraging consistency with} the governing dynamics without { explicit time differentiation}. This is particularly important in higher-order networked systems, where  errors in derivative estimation can be significantly amplified by nonlinear interaction terms, leading to instability in structure identification.

\begin{figure}[t]
    \centering
\includegraphics[width=\linewidth]{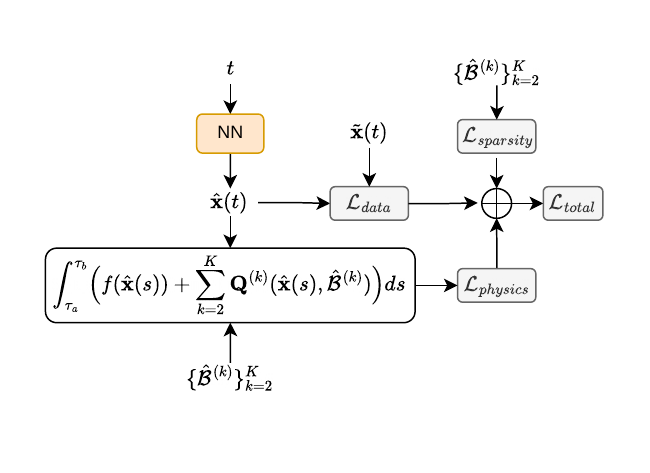}
\vspace{-40pt}
   \caption{Architecture of the proposed IntegralPINN framework. A neural network parameterizes the latent state trajectory from time inputs. The data loss and integral-based physics loss are jointly minimized to learn both the system dynamics and the higher-order interaction tensors.}
    \label{fig:PINN}
\end{figure}

The IntegralPINN framework consists of three main components, as illustrated in Fig.~\ref{fig:PINN}. A neural surrogate is first introduced to approximate the latent continuous-time state trajectory from time inputs. The unknown higher-order interaction tensors are then parameterized as continuous learnable variables, enabling the interaction structure to be inferred jointly with the system trajectory via gradient-based optimization. Finally, physical consistency is { encouraged} through an integral { residual} over sampled time intervals, which improves robustness to noise and naturally accommodates nonsmooth or event-driven dynamics. In particular, the estimated trajectory is used to construct both the data-fitting loss and the integral-based physics loss, which are jointly minimized to learn the system dynamics and higher-order interaction structure.
In the following, we describe the state and interaction parameterization, the integral physics { residual}, and extensions to binary and nonsmooth dynamics.

\subsection{State and Interaction Parameterization}
We first introduce the parameterizations of the system trajectory and the higher-order interaction structure. Specifically, the system state is approximated by a neural network surrogate $\hat{\textbf{x}}(t)$, which maps the scalar time input $t$ to the corresponding state vector. In implementation, we employ a residual multilayer perceptron with fully connected layers to improve approximation accuracy and training stability. Given the discrete observations $\tilde{\textbf{X}}$, the neural surrogate is trained by minimizing the empirical data-fitting loss
\begin{equation*}
\mathcal{L}_{\mathrm{data}}
= \frac{1}{M}\sum_{l=1}^M \|\hat{\textbf{x}}(t_l)-\tilde{\textbf{x}}(t_l)\|_2^2.
\end{equation*}
This parameterization provides a continuous approximation of the latent state trajectory, while implicitly exhibiting a smoothing effect that attenuates high-frequency components induced by measurement noise. Such behavior is consistent with the spectral bias of neural networks, where low-frequency components are learned preferentially over high-frequency ones. This property is essential for stabilizing the subsequent identification of higher-order interactions.

To make this combinatorial reconstruction problem amenable to gradient-based optimization, we introduce a family of trainable tensors $\{\mathscr W^{(k)}\}_{k=2}^K$ and define the estimated interaction tensors $\widehat{\mathscr{B}}^{(k)}$ via the bounded parameterization
\begin{equation*}
\widehat{\mathscr{B}}^{(k)}_{j_1j_2 \cdots j_{k}}
=
{ \mathrm{Sigmoid}}\left(\mathrm{clip}\left(\mathscr W^{(k)}_{j_1j_2 \cdots j_{k}},\, { -c_{\mathrm{clip}}},\, { c_{\mathrm{clip}}}\right)\right),
\end{equation*}
where $\mathrm{clip}(\cdot)$ is applied elementwise to enforce bounded pre-activations, { and $c_{\mathrm{clip}}>0$ is a prescribed clipping threshold that remains fixed during training and is enforced by projecting the pre-activation parameters onto $[-c_{\mathrm{clip}},c_{\mathrm{clip}}]$ after each optimizer update.} This design prevents gradient saturation by keeping the parameters within the sensitive region of the sigmoid during optimization.
Under this parameterization, each entry of $\widehat{\mathscr{B}}^{(k)}$ lies in $(0,1)$ and can be interpreted as a continuous measure of the presence of the corresponding $k$th-order interaction. The tensor $\widehat{\mathscr{B}}^{(k)}$ thus provides a continuous relaxation of the underlying higher-order interaction structure, from which the discrete topology can be recovered via thresholding or sparsity patterns after training. Substituting $\widehat{\mathscr{B}}^{(k)}$ into the governing dynamics yields a fully differentiable model that can be optimized jointly with the neural state surrogate.

\subsection{Integral Physics { Residual}}\label{sec:residual}
To { encourage} consistency with the governing dynamics while avoiding the instability of derivative-based { residuals}, we reformulate the system in an integral form over a time interval $[\tau_a,\tau_b]$. Since the state trajectory satisfies the differential equation, it equivalently obeys an integral relation over any interval. This motivates { penalizing} an integral residual rather than pointwise differential residuals as in standard PINNs. Specifically, for any $\tau_a < \tau_b$, we define the integral residual as
\begin{align*}
\textbf{R}_{ab}
&=
\hat{\textbf{x}}(\tau_b)
-
\hat{\textbf{x}}(\tau_a)
-\\
&\int_{\tau_a}^{\tau_b}
\left(
f(\hat{\textbf{x}}(s))
+
\sum_{k=2}^K \textbf{Q}^{(k)}(\hat{\textbf{x}}(s),\widehat{\mathscr{B}}^{(k)})
\right)ds,
\end{align*}
where the $i$th component of $\textbf{Q}^{(k)}(\hat{\textbf{x}}(s),\widehat{\mathscr{B}}^{(k)})$ is given by
\begin{equation*}
\sum_{j_2=1}^n\sum_{j_3=1}^n\cdots \sum_{j_k=1}^n
\widehat{\mathscr{B}}^{(k)}_{i j_2\cdots j_k}f^{(k)}\big(\hat{x}_i(s),\hat{x}_{j_2}(s),\dots,\hat{x}_{j_k}(s)\big).
\end{equation*}
Ideally, the true trajectory satisfies $\textbf{R}_{ab}=\textbf{0}$. Compared with differential formulations, the integral { residual} avoids explicit differentiation and reduces sensitivity to high-frequency perturbations. Based on the residual, we define the physics-informed loss as
$
\mathcal{L}_{\mathrm{physics}}
=\mathbb{E}_{a,b}\left(\|\textbf{R}_{ab}\|_2^2\right),
$
{ where $\mathbb E_{a,b}(\cdot)$ denotes the expectation over sampled time intervals $[\tau_a,\tau_b]\subset[t_1,t_M]$, approximated by mini-batch Monte Carlo sampling. Specifically, the interval length is sampled uniformly from $(0,h]$, followed by a uniformly sampled starting point within the admissible range, where $h$ is a prescribed maximum window length. The integral is approximated using five-point Gauss--Legendre quadrature evaluated along the neural surrogate  $\hat{\textbf x}$.} The physics loss penalizes violations of the integral form of the governing dynamics over sampled intervals.
{ To promote sparse interaction structures, we further introduce the sparsity penalty
$
\mathcal{L}_{\mathrm{sparsity}}
=
\sum_{k=2}^{K}\big\|\widehat{\mathscr{B}}^{(k)}\big\|_1,
$
where $\|\cdot\|_1$ denotes the entrywise $\ell_1$ norm.}
The neural surrogate and the interaction tensors are then learned jointly by minimizing
\begin{equation*}
\mathcal{L}_{\mathrm{total}}
=
\mathcal{L}_{\mathrm{data}}
+\lambda
\mathcal{L}_{\mathrm{physics}}
{ +\eta\mathcal{L}_{\mathrm{sparsity}}},
\end{equation*}
where $\lambda>0$ { and $\eta>0$ weight physical consistency and sparsity regularization, respectively}.

To improve training stability, we adopt a staged optimization strategy. 
Specifically, the neural surrogate is first trained using only the data-fitting loss $\mathcal{L}_{\mathrm{data}}$ to obtain a reliable approximation of the system trajectory. The integral physics loss $\mathcal{L}_{\mathrm{physics}}$ is then gradually introduced via a ramp-up schedule on $\lambda$, enabling stable joint optimization of the state and interaction parameters. The overall optimization is carried out using stochastic gradient-based methods (e.g., Adam). 

\subsection{Extension to Binary and Non-Smooth Dynamics}\label{sec:binary}
An important advantage of the proposed integral formulation is its ability to accommodate non-smooth and event-driven dynamics that are challenging for standard PINNs. Standard PINNs { penalize residuals} of the governing equations in differential form and therefore rely on pointwise derivatives of the state trajectory. For systems with discontinuities or instantaneous jumps, such derivatives may be undefined or numerically unstable, which can significantly degrade training performance. In contrast, IntegralPINN { encourages} physical consistency over time intervals through an integral { residual}. This replaces pointwise derivative evaluations with finite state increments, thereby naturally absorbing isolated discontinuities and avoiding the need to evaluate derivatives at jump points.

This property is particularly beneficial for binary-state dynamical systems arising in spreading and contagion processes, where $x_i(t)\in\{0,1\}$ and state transitions occur instantaneously. In such settings, the classical derivative $\dot{x}_i(t)$ is not well defined at switching times. However, for any interval $[\tau_a,\tau_b]$, the state increment
$\Delta \textbf{x}=\textbf{x}(\tau_b)-\textbf{x}(\tau_a)$
remains well defined, with each entry taking values in $\{-1,0,1\}$. Consequently, the same interval-based physics { residual} can still be { constructed} through suitable continuous state estimates, allowing the framework to remain applicable under binary observations and discontinuous transitions. { Specifically, for binary observations $\tilde x_i(t_l)\in\{0,1\}$, the surrogate $\hat x_i(t)\in(0,1)$ obtained through a sigmoid output layer is interpreted as the probability of the binary state. Accordingly, the data-fitting loss is defined using binary cross-entropy
\begin{equation*}
\mathcal L_{\mathrm{data}}
=
\frac{1}{nM}\sum_{l=1}^{M}\sum_{i=1}^{n}
\ell_{\mathrm{BCE}}\big(\hat x_i(t_l),\tilde x_i(t_l)\big),
\end{equation*}
where $\ell_{\mathrm{BCE}}(p,y)=-y\log p-(1-y)\log(1-p)$.
The integral residual $\textbf R_{ab}$ is then evaluated along this surrogate using $\hat{\textbf x}(\tau_a)$, $\hat{\textbf x}(\tau_b)$ and the model dynamics along $\hat{\textbf x}(t)$.} In addition to binary spreading processes, the same formulation readily extends to other non-smooth systems, including impulsive dynamics and switched interaction mechanisms, where piecewise integral { residuals} remain well defined even when derivatives do not.

\section{Robustness Analysis}\label{sec:theory}
For theoretical analysis, we rewrite the governing higher-order interaction dynamics  in the compact parameterized form
$
\dot{\textbf{x}}=F(\textbf{x})\textbf b,
$
where $F(\textbf{x})$ denotes the dictionary matrix induced by the local dynamics and higher-order interaction functions, and $\textbf{b}$ is the stacked interaction parameter vector containing all entries of the interaction tensors $\{\mathscr{B}^{(k)}\}_{k=2}^K$. This representation is algebraically equivalent to the original model and facilitates the comparison between derivative-based and integral-based physics { residuals}. Let $\textbf{b}^\star$ denote the true interaction parameter vector. Given  noisy discrete measurements $\tilde{\textbf{X}}$, a continuous neural surrogate $\hat{\textbf{x}}(t)$ is trained to fit the observations{, and the physics residuals are evaluated using this surrogate, with time derivatives computed by differentiating the surrogate}.

{ We compare the two residual formulations through their corresponding least-squares estimators  $\hat{\textbf b}_{\mathrm{diff}}$ and $\hat{\textbf b}_{\mathrm{int}}$, in which the surrogate $\hat{\textbf x}$ and the window length $h$ are fixed and $\textbf b$ minimizes the physics residual alone. These estimators characterize the parameter values favored by each residual formulation under the same surrogate error, rather than the outcome of joint neural network and parameter optimization. In the implemented IntegralPINN, the bounded sigmoid parameterization generally excludes $\textbf b^\star$ from the feasible set, so the learned coefficients are interpreted as interaction scores rather than estimates of $\textbf b^\star$. The analysis therefore isolates the effects of the residual formulations, without accounting for the bounded parameterization, sparsity regularization, random window lengths, or joint optimization.

We consider a finite horizon $[0,T]$ on which the true solution exists. Since $\textbf x(t)$ and $\hat{\textbf x}(t)$ are continuous on $[0,T]$, they are bounded and lie in a compact ball $\mathcal D_R=\{\textbf z\in\mathbb R^n:\|\textbf z\|_2\leq R\}$ for some constant $R>0$. We  assume that $F$ is continuously differentiable on an open neighborhood of $\mathcal D_R$, and hence uniformly bounded and Lipschitz continuous on $\mathcal D_R$, with constants $C_F$ and $L_F$, respectively.}
We focus on the practically relevant regime in which the learned surrogate provides accurate trajectory fitting, while the remaining residual may contain small high-frequency artifacts.

\begin{assumption}\label{ass:bound}
Let $\boldsymbol{\delta}(t)=\hat{\textbf{x}}(t)-\textbf{x}(t)$ denote the surrogate approximation error. { Assume that} there exist two constants $C_0,C_1>0$ such that
\[
\|\boldsymbol{\delta}\|_{2} \le C_0 \sigma,\quad
\|\dot{\boldsymbol{\delta}}\|_{2} \le C_1 \frac{\sigma}{\tau},
\]
where $\|\cdot\|_2$ denotes the $L_2$ norm over the observation horizon, $\sigma$ denotes the observation noise level, { and $\tau>0$ denotes a characteristic time scale of the residual oscillatory component, with smaller $\tau$ corresponding to higher-frequency residual variations.}
\end{assumption}

{ The second bound characterizes a worst-case scenario in which the remaining surrogate error contains a high-frequency component with amplitude of order $\mathcal{O}(\sigma)$. For example, a residual component $\boldsymbol{\delta}(t)=\sigma \textbf{a}\sin(t/\tau)$ satisfies $\dot{\boldsymbol{\delta}}(t)=(\sigma/\tau)\textbf{a}\cos(t/\tau)$, which directly illustrates the $\mathcal{O}(\sigma/\tau)$ scaling. The spectral bias of neural networks may attenuate such high-frequency components, but does not guarantee their complete elimination from the remaining surrogate error. In practice, such residual components may arise when the surrogate is trained on observations corrupted by broadband noise or when the underlying dynamics contain fast time scales or sharp transitions that are difficult to approximate accurately.}

\subsection{Differential Formulation}
We first examine the derivative-based physics { residual} employed in standard PINNs, where consistency with the governing dynamics is { promoted} through  time derivatives. For theoretical comparison, suppose that a surrogate trajectory $\hat{\textbf{x}}(t)$ has been obtained. The estimated interaction parameter vector induced by differential residual minimization is given by
\begin{equation*}
\hat{\textbf b}_{\mathrm{diff}}
=
\arg\min_{\textbf b}
\int_0^T
\|\dot{\hat{\textbf{x}}}(t)-F(\hat{\textbf{x}}(t))\textbf b\|_2^2\,dt.
\end{equation*}
\begin{assumption}\label{ass:excitation}
There exists a constant $\lambda_0>0$ such that
\begin{equation*}
\lambda_{\min}\left(
\int_0^T
F({ \textbf{x}}(t))^\top
F({ \textbf{x}}(t))dt\right)\ge \lambda_0.
\end{equation*}
\vspace{-20pt}
\end{assumption}
This condition ensures that the { regression matrix induced by the true trajectory} remains sufficiently informative over the observation horizon. In particular, the associated Gram matrix is uniformly nonsingular, and therefore
\begin{equation*}
\left\|
\left(
\int_0^T
F({ \textbf{x}}(t))^\top
F({ \textbf{x}}(t))
dt
\right)^{-1}
\right\|_2
\le \frac{1}{\lambda_0}.
\end{equation*}

\begin{theorem}\label{thm:derivative}
Suppose that Assumptions~\ref{ass:bound} and~\ref{ass:excitation} hold. Then, { for sufficiently small $\sigma$,} the estimator $\hat{\textbf b}_{\mathrm{diff}}$ satisfies
\begin{equation*}
\|\hat{\textbf b}_{\mathrm{diff}}-\textbf b^\star\|_2
=
\mathcal O(\sigma)
+
\mathcal O\left(\frac{\sigma}{\tau}\right).
\end{equation*}
\vspace{-20pt}
\end{theorem}
\textbf{Proof.}
Since $\hat{\textbf{x}}=\textbf{x}+\boldsymbol{\delta}$, we have
$
\dot{\hat{\textbf{x}}}
=\dot{\textbf{x}}+\dot{\boldsymbol{\delta}}=F(\textbf{x})\textbf b^\star+\dot{\boldsymbol{\delta}}.
$
Define $\Delta F=F(\hat{\textbf{x}})-F(\textbf{x}).$
Then
$\dot{\hat{\textbf{x}}}
=F(\hat{\textbf{x}})\textbf b^\star+\boldsymbol{\varepsilon}_{\mathrm{diff}},$
where
$\boldsymbol{\varepsilon}_{\mathrm{diff}}
=\dot{\boldsymbol{\delta}}-\Delta F\textbf b^\star.$
Since the objective is quadratic in $\textbf b$, the minimizer satisfies the normal equation
\begin{equation*}
{ G_{\hat{x}}(\hat{\textbf b}_{\mathrm{diff}}-\textbf b^\star)}
=
\int_0^T
F(\hat{\textbf{x}}(t))^\top
\boldsymbol{\varepsilon}_{\mathrm{diff}}(t)dt,
\end{equation*}
{ where $G_{\hat{x}}=\int_0^T F(\hat{\textbf{x}}(t))^\top F(\hat{\textbf{x}}(t))dt$. To relate the excitation condition on the true trajectory to that of the learned surrogate, let $G_x=\int_0^T F(\textbf{x}(t))^\top F(\textbf{x}(t))dt$. Since $F(\hat{\textbf{x}})=F(\textbf{x})+\Delta F$, we have
\begin{equation*}
\scalebox{0.92}{$\displaystyle
G_{\hat{x}}-G_x
=
\int_0^T
\left[
F(\textbf{x})^\top\Delta F
+
\Delta F^\top F(\textbf{x})
+
\Delta F^\top\Delta F
\right]dt.$}
\end{equation*}}
Since $F(\textbf{x})$ is Lipschitz continuous, there exists $L_F>0$ such that $\|\Delta F(t)\|_2\le L_F\|\boldsymbol{\delta}(t)\|_2$, $t\in[0,T]$. { Consequently, by Assumption~\ref{ass:bound}, $\|\Delta F\|_2\le L_F\|\boldsymbol{\delta}\|_2\le L_F C_0\sigma$. Moreover,} since $F(\textbf{x})$ is uniformly bounded on the compact domain{\ containing both $\textbf{x}$ and $\hat{\textbf{x}}$}, there exists a constant $C_F>0$ such that { $\|F(\textbf{x}(t))\|_2,\,$}$\|F(\hat{\textbf{x}}(t))\|_2\le C_F$, $t\in[0,T]$. { Therefore, by the Cauchy--Schwarz inequality,
\begin{equation*}
\|G_{\hat{x}}-G_x\|_2
\le
2C_F\sqrt{T}\|\Delta F\|_2
+
\|\Delta F\|_2^2
=
\mathcal O(\sigma).
\end{equation*}
By Weyl's inequality and Assumption~\ref{ass:excitation},
\begin{equation*}
\lambda_{\min}(G_{\hat{x}})
\ge
\lambda_{\min}(G_x)
-
\|G_{\hat{x}}-G_x\|_2
\ge
\lambda_0-\mathcal O(\sigma).
\end{equation*}
Hence, for sufficiently small $\sigma$, $G_{\hat{x}}$ remains nonsingular and $\|G_{\hat{x}}^{-1}\|_2\le 1/(\lambda_0-\mathcal O(\sigma))=\mathcal O(1)$.}
Taking norms { in the normal equation after multiplying by $G_{\hat{x}}^{-1}$} and using { the above bound} yields
\begin{equation*}
\|\hat{\textbf b}_{\mathrm{diff}}-\textbf b^\star\|_2
\le
{\mathcal O(1)}
\left\|
\int_0^T
F(\hat{\textbf{x}}(t))^\top
\boldsymbol{\varepsilon}_{\mathrm{diff}}(t)dt\right\|_2.
\end{equation*}
Hence, by the Cauchy--Schwarz inequality,
\begin{equation*}
\left\|
\int_0^T
F(\hat{\textbf{x}}(t))^\top
\boldsymbol{\varepsilon}_{\mathrm{diff}}(t)dt
\right\|_2\le
C_F\sqrt{T}\,
\|\boldsymbol{\varepsilon}_{\mathrm{diff}}\|_2.
\end{equation*}
Therefore, $\|\hat{\textbf b}_{\mathrm{diff}}-\textbf b^\star\|_2
\le{\mathcal O(1)}
\|\boldsymbol{\varepsilon}_{\mathrm{diff}}\|_2.$
Using the triangle inequality,
$\|\boldsymbol{\varepsilon}_{\mathrm{diff}}\|_2
\le
\|\dot{\boldsymbol{\delta}}\|_2+
\|\Delta F\,\textbf b^\star\|_2.$
Thus, $\|\Delta F\,\textbf b^\star\|_2
\le
L_F\|\textbf b^\star\|_2
\|\boldsymbol{\delta}\|_2
=\mathcal O(\sigma).$
Moreover, by Assumption~\ref{ass:bound}, it holds that
$\|\dot{\boldsymbol{\delta}}\|_2
=
\mathcal O(\sigma/\tau).$
Combining the above bounds yields $\|\hat{\textbf b}_{\mathrm{diff}}-\textbf b^\star\|_2=\mathcal O(\sigma)+\mathcal O(\sigma/\tau)$, which establishes the stated result.\hfill $\blacksquare$

Theorem~\ref{thm:derivative} shows that, besides the baseline approximation error of order $\mathcal O(\sigma)$, the derivative-based formulation introduces an additional amplification term of order $\mathcal O(\sigma/\tau)$ caused by differentiation of residual high-frequency artifacts.

\subsection{Integral Formulation}
We next consider the integral formulation underlying the proposed IntegralPINN framework. For theoretical comparison, we restrict the integration interval to a moving window of fixed length $h$, namely $\tau_a=t$ and $\tau_b=t+h$. 
The estimated interaction parameter vector induced by integral residual minimization is defined as
\begin{equation*}
\scalebox{0.87}{$\displaystyle
\hat{\textbf b}_{\mathrm{int}}
=
\arg\min_{\textbf b}
\int_0^{T-h}
\left\|
\hat{\textbf{x}}(t+h)-\hat{\textbf{x}}(t)
-
\int_t^{t+h}F(\hat{\textbf{x}}(s))\textbf b ds
\right\|_2^2 dt$.}
\end{equation*}

\begin{assumption}\label{ass:int_excitation}
{ For a fixed window length $h\in(0,T)$, assume that there exists $\mu_0(h)>0$ such that
\begin{equation*}
\lambda_{\min}\left(
\int_0^{T-h}
\mathcal I_{F,x}(t)^\top\mathcal I_{F,x}(t)\,dt
\right)
\geq \mu_0(h),
\end{equation*}
where $\mathcal I_{F,x}(t)=\int_t^{t+h}F(\textbf{x}(s))\,ds$ denotes the integrated regression matrix associated with the true trajectory over the window $[t,t+h]$.}
\end{assumption}

\begin{theorem}\label{thm:integral}
{ Suppose that Assumptions~\ref{ass:bound} and~\ref{ass:int_excitation} hold. Then, for any fixed $h\in(0,T)$ and sufficiently small $\sigma$,
\begin{equation*}
\|\hat{\textbf{b}}_{\mathrm{int}}-\textbf{b}^\star\|_2
=
\mathcal O\left(
\frac{h}{\mu_0(h)}
\left(
\min\left\{1,\frac{h}{\tau}\right\}+h
\right)\sigma
\right),
\end{equation*}
where the hidden constants are independent of $\sigma$, $h$, and $\tau$, while the admissible noise level may depend on $h$.}
\end{theorem}
\textbf{Proof.}
Using $\hat{\textbf{x}}=\textbf{x}+\boldsymbol{\delta}$ and the true dynamics $\dot{\textbf{x}}=F(\textbf{x})\textbf b^\star${,} integration over $[t,t+h]$ gives
\begin{equation*}
\hat{\textbf{x}}(t+h)-\hat{\textbf{x}}(t)
=
\int_t^{t+h}F(\hat{\textbf{x}}(s))\textbf b^\star ds
+
\boldsymbol{\varepsilon}_{\mathrm{int}}(t),
\end{equation*}
where
$\boldsymbol{\varepsilon}_{\mathrm{int}}(t)
=
\boldsymbol{\delta}(t+h)-\boldsymbol{\delta}(t)
-
\int_t^{t+h}\Delta F(s)\textbf b^\star ds$ and $\Delta F(s)=F(\hat{\textbf{x}}(s))-F(\textbf{x}(s))$. Since the objective is quadratic in $\textbf b$, the minimizer satisfies 
\begin{equation*}
{ G_{h,\hat{x}}(\hat{\textbf b}_{\mathrm{int}}-\textbf b^\star)}
=
\int_0^{T-h}\mathcal I_{F,\hat{x}}(t)^\top\boldsymbol{\varepsilon}_{\mathrm{int}}(t)\,dt,
\end{equation*}
where $G_{h,\hat{x}}=\int_0^{T-h}\mathcal I_{F,\hat{x}}(t)^\top\mathcal I_{F,\hat{x}}(t)dt$ { and $\mathcal I_{F,\hat{x}}(t)=\int_t^{t+h}F(\hat{\textbf{x}}(s))\,ds$}. Since $F$ is uniformly bounded on the compact domain, there exists $C_F>0$ such that $\|\mathcal I_{F,\hat{x}}(t)\|_2\le hC_F$ { and $\|\mathcal I_{F,x}(t)\|_2\le hC_F$}.
{ We first transfer the excitation condition from the true trajectory to the learned surrogate. Let $$\Delta\mathcal I_F(t)=\mathcal I_{F,\hat{x}}(t)-\mathcal I_{F,x}(t)=\int_t^{t+h}\Delta F(s)ds.$$ By the Lipschitz continuity of $F$ and Assumption~\ref{ass:bound}, $\|\Delta\mathcal I_F\|_2\leq hL_F\|\boldsymbol{\delta}\|_2\leq hL_FC_0\sigma$. Let $$G_{h,x}=\int_0^{T-h}\mathcal I_{F,x}(t)^\top\mathcal I_{F,x}(t)\,dt.$$ Then
$\|G_{h,\hat{x}}-G_{h,x}\|_2
\leq
h^2L_FC_0\sigma\bigl(2C_F\sqrt{T-h}+L_FC_0\sigma\bigr),$
which we denote by $\xi(h,\sigma)$. By Weyl's inequality and Assumption~\ref{ass:int_excitation}, $\lambda_{\min}(G_{h,\hat{x}})\geq\mu_0(h)-\xi(h,\sigma)$. For fixed $h$, sufficiently small $\sigma$ ensures $\xi(h,\sigma)\leq\mu_0(h)/2$, and hence $\|G_{h,\hat{x}}^{-1}\|_2\leq 2/\mu_0(h)$.} Hence,
\[
\|\hat{\textbf b}_{\mathrm{int}}-\textbf b^\star\|_2
\le
\frac{{ 2}hC_F\sqrt{T-h}}{{ \mu_0(h)}}
\|\boldsymbol{\varepsilon}_{\mathrm{int}}\|_{2}.
\]
By the triangle inequality, it holds that
\[
\|\boldsymbol{\varepsilon}_{\mathrm{int}}\|_{2}
\le
\|\boldsymbol{\delta}(t+h)-\boldsymbol{\delta}(t)\|_{2}
+
\left\|
\int_t^{t+h}\Delta F(s)\textbf b^\star ds
\right\|_{2}.
\]
For the first term, { $\|\boldsymbol{\delta}(t+h)-\boldsymbol{\delta}(t)\|_2\le 2C_0\sigma$. On the other hand, $\boldsymbol{\delta}(t+h)-\boldsymbol{\delta}(t)=\int_t^{t+h}\dot{\boldsymbol{\delta}}(s)\,ds$ gives $\|\boldsymbol{\delta}(t+h)-\boldsymbol{\delta}(t)\|_2\le hC_1\sigma/\tau$. Thus,
$
\|\boldsymbol{\delta}(t+h)-\boldsymbol{\delta}(t)\|_2
\le
\min\{2C_0,\frac{hC_1}{\tau}\}\sigma.
$}
For the second term, the Lipschitz continuity of $F$ implies $\|\Delta F(s)\|_2
\le
L_F\|\boldsymbol{\delta}(s)\|_2.$
Hence,
\begin{equation*}
\left\|
\int_t^{t+h}\Delta F(s)\textbf b^\star\,ds
\right\|_2
\le
h L_F \|\textbf b^\star\|_2
{ C_0\sigma}.
\end{equation*}
Thus,
{
\[
\|\boldsymbol{\varepsilon}_{\mathrm{int}}\|_{2}
\le
\left(
\min\left\{2C_0,\frac{hC_1}{\tau}\right\}
+
hL_FC_0\|\textbf{b}^\star\|_2
\right)\sigma,\]} 
which yields { the bound stated in Theorem~\ref{thm:integral}} and establishes the stated result. \hfill $\blacksquare$

{ Theorem~\ref{thm:integral} assumes exact integration. Numerical quadrature perturbs both the integrated regression matrix and the residual. As long as this perturbation is small compared with $\mu_0(h)$, it adds a quadrature-dependent term to the bound, which tends to zero as the number of quadrature nodes increases since the integrand is continuous on each interval.}

\subsection{Frequency-Domain Interpretation}\label{sec:freq}

The preceding time-domain analysis shows that the integral formulation { mitigates derivative-induced error amplification for fixed integration windows}. We next provide a complementary frequency-domain interpretation by examining how differentiation and integration affect high-frequency components of the surrogate approximation error. To this end, we consider the fixed-window interval residual and isolate the contribution of the surrogate error. Let $\widehat{\boldsymbol{\delta}}(\omega)$ denote the Fourier transform of the surrogate error $\boldsymbol{\delta}(t)$. 
Differentiation corresponds in the frequency domain to multiplication by $i\omega$, i.e.,
$\mathcal F\{\dot{\boldsymbol{\delta}}\}(\omega)
=
i\omega\,\widehat{\boldsymbol{\delta}}(\omega),$
and therefore amplifies high-frequency errors. 
By contrast, define the fixed-window integral operator acting on a generic signal $g$ as
$(\mathcal I_h g)(t)=\int_t^{t+h} g(s)ds.$
Up to a time-shift convention, this operator corresponds to convolution with the box kernel $k_h(t) = \mathbf{1}_{[0,h]}(t)$. Its frequency response is
$\widehat{k}_h(\omega)=(e^{i\omega h}-1)/(i\omega)$, which satisfies $|\widehat{k}_h(\omega)|=\mathcal{O}(|\omega|^{-1})$ for $|\omega|\to\infty.$
Thus, the integral operator attenuates high-frequency components.

Note that 
\begin{equation*}
\boldsymbol{\delta}(t+h)-\boldsymbol{\delta}(t)=\int_t^{t+h}\dot{\boldsymbol{\delta}}(s)\,ds=(\mathcal I_h\dot{\boldsymbol{\delta}})(t).    
\end{equation*}
Applying the integral operator after differentiation yields
$
\mathcal F\{\mathcal I_h\dot{\boldsymbol{\delta}}\}(\omega)
=
\widehat{k}_h(\omega)\,i\omega\,\widehat{\boldsymbol{\delta}}(\omega)
=
\left(e^{i\omega h}-1\right)\widehat{\boldsymbol{\delta}}(\omega).
$
Thus, the unbounded factor $i\omega$ introduced by differentiation is canceled by the $1/(i\omega)$ decay of the integral kernel, and the resulting frequency multiplier { satisfies $|e^{i\omega h}-1|\leq 2$. For small windows, however, $e^{i\omega h}-1\approx i\omega h$, indicating that the mitigation becomes significant only at frequencies on the order of $1/h$ or higher. Consistently, as  $h\rightarrow 0$, the Gram matrix in Assumption~\ref{ass:int_excitation} is approximately $h^2$ times that in Assumption~\ref{ass:excitation}. Therefore, $\mu_0(h)$ can be chosen proportional to $\lambda_0h^2$, and Theorem~\ref{thm:integral} recovers the $\mathcal O(\sigma)+\mathcal O(\sigma/\tau)$ scaling of Theorem~\ref{thm:derivative}. The bound in Theorem~\ref{thm:integral} avoids the derivative-amplification factor when $h\geq 2C_0\tau/C_1$, whereas excessively long windows may reduce $\mu_0(h)$. Hence, the integral formulation does not uniformly outperform the differential formulation, and the window length should be chosen accordingly.}

\section{Numerical Experiments}\label{sec:example}
We evaluate the proposed IntegralPINN on three representative scenarios and compare its performance against recent  state-of-the-art methods including  HyperPINN (2025) \cite{mao2025identification}, the kernel-based method (2026) \cite{donati2026kernel},  THIS  (2025) (THIS does not require prior knowledge of the coupling functions) \cite{delabays2025hypergraph}{ , and integral SINDy (2017) \cite{schaeffer2017sparse}}. The selected scenarios are designed to highlight complementary challenges: 
(i) dense nonlinear dynamics with continuous observations, 
(ii) discontinuous binary observations arising from complex contagion processes, and 
(iii) moderately sized neuronal systems with heterogeneous interaction structures. { All simulations were conducted in Python 3 on NVIDIA RTX A6000 GPUs.} The code for reproducing these experiments is available at \url{https://github.com/shahinxu/Integral_PINN.git}.

{ Reconstruction performance is evaluated using the area under the receiver operating characteristic curve (AUROC), computed separately for each interaction order $k$ over candidate node sets of size $k$, with those in the ground truth labeled positive. The coefficient associated with each node set serves as its score, and the estimated weights of the baselines are used analogously. The Macro-AUC is defined as the unweighted mean of the per-order AUROC values. The results  are reported as mean $\pm$ standard deviation together with 95\% confidence intervals based on the $t$-distribution over 10 random seeds, which vary the network initialization, the interval sampling, and the observation noise.
Unless otherwise stated, Gaussian noise with a relative level of \(\sigma=0.02\) is added to continuous-valued observations, while binary observations are independently flipped with probability \(\sigma=0.02\).
 Due to computational constraints, all methods are evaluated using their  default hyperparameters (see Table A1 in the Appendix). }

\begin{table*}[t]

\setlength{\tabcolsep}{4pt}
\caption{AUROC of higher-order interaction reconstruction for ecological networks (maximum interaction order $K=5$). $k$ denotes the interaction order evaluated in each column. The statistical significance of Macro-AUC differences between IntegralPINN and each baseline is reported in Table A2 of the Appendix.}
\label{tab:scenario1}
\centering
\begin{tabular}{@{}l cccccc@{}}
\hline
Method & $k=2$ & $k=3$ & $k=4$ & $k=5$ & Macro-AUC & 95\% CI \\
\hline
IntegralPINN & $0.835\pm0.022$ & $0.751\pm0.025$ & $0.718\pm0.028$ & $0.675\pm0.033$ & $0.745\pm0.023$ & [0.729, 0.761] \\
HyperPINN & $0.619\pm0.048$ & $0.563\pm0.079$ & $0.603\pm0.045$ & $0.717\pm0.173$ & $0.625\pm0.069$ & [0.576, 0.675] \\
Kernel-based & $0.438\pm0.082$ & $0.521\pm0.068$ & $0.548\pm0.067$ & $0.548\pm0.146$ & $0.514\pm0.043$ & [0.483, 0.545] \\
THIS & $0.541\pm0.108$ & $0.497\pm0.072$ & $0.512\pm0.130$ & $0.611\pm0.174$ & $0.540\pm0.060$ & [0.498, 0.583] \\
Integral SINDy & $0.435\pm0.088$ & $0.500\pm0.047$ & $0.493\pm0.038$ & $0.490\pm0.158$ & $0.479\pm0.054$ & [0.441, 0.518] \\
\hline
\end{tabular}
\end{table*}

\subsection{Ecological Population}\label{sec:eco}
We first consider an ecological system with higher-order interactions with $9$ species organized into three trophic levels, consisting of $3$ producers, $3$ primary consumers, and $3$ secondary consumers. The dynamics of each species follows the model in \cite{grilli2017higher} and is described as 
\begin{align*}
&\dot{x}_i
=\,
f(x_i)
+\sum_{j_2=1}^9 \mathscr{B}^{(2)}_{ij_2}
f^{(2)}(x_i,x_{j_2}) \\
&- x_i
\sum_{k=3}^{5}\sum_{j_2=1}^9\cdots\sum_{j_k=1}^9
\mathscr{B}^{(k)}_{i j_2\dots j_{k}}
f^{(k)}(x_i,x_{j_2},\dots,x_{j_{k}}),
\end{align*}
where $f(x_i)$ models logistic growth for producers and natural mortality for consumers, $f^{(2)}(x_i,x_{j_2})$ denotes pairwise predation, and $f^{(k)}(x_i,x_{j_2},\dots,x_{j_{k}})$ for $k=3,4,5$ represents higher-order competitive effects. The full candidate set contains $372$ potential higher-order interactions. The model is trained using $300$ temporal observations. To simulate measurement uncertainty, Gaussian noise with a relative level of $2\%$ is added to the observations.

The performance comparison with the { four} baseline methods is reported in Table~\ref{tab:scenario1}.
The proposed IntegralPINN { achieves the highest Macro-AUC and the highest AUROC for $k=2,3,4$}, improving by approximately { 19\%} over the second-best method { in Macro-AUC} and maintaining stable performance in both lower- and higher-order regimes. { Although HyperPINN attains a higher mean AUROC for $k=5$, it exhibits substantially greater variability across runs.  The kernel-based method and integral SINDy remain close to chance level for most interaction orders.} The THIS method achieves slightly improved results in certain higher-order entries compared to { the kernel-based method and integral SINDy}, but overall remains unstable and does not show consistent gains with increasing interaction order.
These results { suggest} that IntegralPINN effectively mitigates instability caused by derivative-based formulations and { encourages consistency with the} system dynamics through an integral { residual}, leading to more reliable reconstruction of both trophic and higher-order competitive interactions from noisy observations.

\begin{table*}[t]

\setlength{\tabcolsep}{4pt}
\caption{AUROC of higher-order interaction reconstruction for social contagion networks (maximum interaction order $K=5$). $k$ denotes the interaction order evaluated in each column. The statistical significance of Macro-AUC differences between IntegralPINN and each baseline is reported in Table A2 of the Appendix.}
\label{tab:scenario2}
\centering
\begin{tabular}{@{}l cccccc@{}}
\hline
Method & $k=2$ & $k=3$ & $k=4$ & $k=5$ & Macro-AUC & 95\% CI \\
\hline
IntegralPINN & $0.751\pm0.030$ & $0.704\pm0.026$ & $0.785\pm0.029$ & $0.680\pm0.038$ & $0.730\pm0.024$ & [0.713, 0.747] \\
HyperPINN & $0.600\pm0.094$ & $0.679\pm0.083$ & $0.641\pm0.076$ & $0.419\pm0.280$ & $0.585\pm0.074$ & [0.532, 0.638] \\
Kernel-based & $0.471\pm0.111$ & $0.503\pm0.068$ & $0.542\pm0.108$ & $0.475\pm0.249$ & $0.498\pm0.067$ & [0.450, 0.545] \\
THIS & $0.483\pm0.106$ & $0.429\pm0.113$ & $0.427\pm0.074$ & $0.363\pm0.176$ & $0.425\pm0.075$ & [0.372, 0.479] \\
Integral SINDy & $0.462\pm0.042$ & $0.498\pm0.019$ & $0.503\pm0.004$ & $0.501\pm0.003$ & $0.491\pm0.012$ & [0.482, 0.500] \\
\hline
\end{tabular}
\end{table*}

\subsection{Social Contagion}
In the second example, we consider a spreading process for epidemic diffusion and information propagation based on the Susceptible--Infected--Susceptible model. We construct a modular higher-order network with $8$ nodes consisting of three interconnected communities. The dynamics follows the model in \cite{iacopini2019simplicial} described as
\begin{equation*}
\scalebox{0.82}{$
\begin{aligned}
&\dot{x}_i=-\mu x_i+(1-x_i) \\
&\left[
\beta \sum_{j_2=1}^8 \mathscr{B}^{(2)}_{i j_2} x_{j_2}
+\beta_\delta \sum_{k=3}^{5} \sum_{j_2=1}^8\cdots\sum_{j_k=1}^8
\mathscr{B}^{(k)}_{i j_2\dots j_k}
f^{(k)}(x_i,x_{j_2},\dots,x_{j_k})
\right],
\end{aligned}
$}
\end{equation*}
where $x_i\in[0,1]$ denotes the infection probability of node $i$ and $\mu=0.4$ is the recovery rate. To capture complex contagion effects, higher-order group exposure is set to be stronger than pairwise infection, with $\beta_\delta=2.0$ and $\beta=0.5$. The candidate set contains $210$ higher-order interactions, among which only $26$ correspond to true interactions.  Although the governing dynamics evolves in continuous time over $t \in [0,50]$, the available observations consist of 300 binary snapshots generated via Bernoulli sampling. Such observations are discontinuous and highly non-Gaussian, posing challenges for methods that rely on derivative information.  

The performance comparison is reported in Table~\ref{tab:scenario2}. The proposed IntegralPINN consistently achieves the highest AUROC across all interaction orders{ , with a Macro-AUC of 0.730 compared with 0.585 for HyperPINN}. HyperPINN shows moderate performance but degrades as the interaction order increases, while kernel-based methods perform poorly and inconsistently, particularly in higher-order regimes. { THIS and integral SINDy perform close to or below chance level.} These results highlight the effectiveness of the proposed approach in handling sparse and heterogeneous interaction structures. By { encouraging consistency with} the governing dynamics through an integral formulation and incorporating binary observations via a  cross-entropy loss, IntegralPINN enables accurate reconstruction of latent higher-order interactions from sparse and highly non-Gaussian observations.

\begin{table*}[t]

\setlength{\tabcolsep}{4pt}
\caption{AUROC of higher-order interaction reconstruction for neuronal networks (maximum interaction order $K=5$). $k$ denotes the interaction order evaluated in each column. The statistical significance of Macro-AUC differences between IntegralPINN and each baseline is reported in Table A2 of the Appendix.}
\label{tab:scenario3}
\centering
\begin{tabular}{@{}l cccccc@{}}
\hline
Method & $k=2$ & $k=3$ & $k=4$ & $k=5$ & Macro-AUC & 95\% CI \\
\hline
IntegralPINN & $0.708\pm0.016$ & $0.705\pm0.021$ & $0.682\pm0.024$ & $0.641\pm0.030$ & $0.684\pm0.018$ & [0.671, 0.697] \\
HyperPINN & $0.486\pm0.034$ & $0.512\pm0.050$ & $0.509\pm0.051$ & $0.533\pm0.116$ & $0.510\pm0.044$ & [0.479, 0.541] \\
Kernel-based & $0.500\pm0.000$ & $0.518\pm0.053$ & $0.490\pm0.074$ & $0.522\pm0.053$ & $0.508\pm0.032$ & [0.485, 0.530] \\
THIS & $0.422\pm0.016$ & $0.582\pm0.018$ & $0.560\pm0.020$ & $0.548\pm0.019$ & $0.528\pm0.038$ & [0.504, 0.552] \\
Integral SINDy & $0.546\pm0.022$ & $0.501\pm0.039$ & $0.498\pm0.012$ & $0.496\pm0.027$ & $0.510\pm0.016$ & [0.499, 0.522] \\
\hline
\end{tabular}
\end{table*}

\subsection{Neuronal Phase Synchronization}\label{sec:neuronal}

The final example considers a multilayer Kuramoto oscillator network with 30 nodes, modeling the coupling between slow and fast neuronal phases. A key characteristic of this system is the separation of interaction structures across the two phase variables: the fast phase is governed solely by pairwise interactions, whereas the slow phase depends on higher-order simplicial interactions up to order five. The dynamics follows the model in \cite{skardal2019abrupt} and is described as
\begin{equation*}
\scalebox{0.94}{$
\begin{aligned}
\dot{\theta}_i
&=
\omega_i
+
\sum_{k=3}^5 \gamma^{(k)}\sum_{j_2=1}^{30}\cdots\sum_{j_k=1}^{30}
\mathscr{B}^{(k)}_{ij_2\dots j_k} f^{(k)}(\theta_i,\theta_{j_2},\cdots,\theta_{j_k}),
\\
\dot{\phi}_i
&=
\nu_i + d\sin(\theta_i-\phi_i)
+
\kappa \sum_{j_2=1}^{30}\mathscr{B}^{(2)}_{ij_2}\sin(\phi_{j_2}-\phi_i),
\end{aligned}$}
\end{equation*}
where $\omega_i$ and $\nu_i$ denote the natural frequencies, and $d$, $\kappa$, and $\gamma^{(k)}$ are coupling coefficients. This setting poses a challenging reconstruction problem due to the coexistence of heterogeneous interaction mechanisms and rapidly growing combinatorial complexity. The candidate space contains more than $1.7 \times 10^5$ potential higher-order interactions. For computational reasons, all methods are evaluated on a candidate set comprising all true interactions and 3,000 randomly sampled non-interacting candidates per order. The reconstruction performance is summarized in Table~\ref{tab:scenario3}, showing consistent trends with those observed in the previous experiments. These results further demonstrate that IntegralPINN can effectively handle heterogeneous coupling structures in medium-scale systems, while maintaining robustness under noisy observations.

{ \subsection{Ablation and Sensitivity Analysis}
\label{sec:ablation}
To assess the contributions of individual components, we perform a controlled ablation experiment over three random seeds by replacing the integral residual with a derivative residual, separately removing the $\lambda$ schedule, bounded tensor parameterization, or pre-activation clipping, and comparing joint optimization with a two-stage variant that freezes the surrogate after data fitting. As shown in Table~\ref{tab:ablation}, IntegralPINN achieves higher Macro-AUC than all these variants in the three tested systems. The integral residual contributes most in the social and neuronal examples, whereas the staged schedule and bounded parameterization have the largest effects in the ecological example.

\begin{table}[t]

\setlength{\tabcolsep}{3pt}
\caption{Ablation study of the proposed IntegralPINN framework.}
\label{tab:ablation}
\centering
\resizebox{\ifdim\width>\linewidth\linewidth\else\width\fi}{!}{%
\begin{tabular}{@{}l ccc@{}}
\hline
Variant & Ecological & Social & Neuronal \\
\hline
IntegralPINN & $0.747\pm0.029$ & $0.727\pm0.022$ & $0.683\pm0.016$ \\
Derivative residual & $0.684\pm0.021$ & $0.547\pm0.012$ & $0.531\pm0.028$ \\
No $\lambda$ schedule & $0.622\pm0.095$ & $0.628\pm0.037$ & $0.601\pm0.062$ \\
Unbounded & $0.648\pm0.103$ & $0.649\pm0.042$ & $0.629\pm0.039$ \\
No clipping & $0.671\pm0.044$ & $0.681\pm0.069$ & $0.642\pm0.051$ \\
Two-stage & $0.688\pm0.021$ & $0.644\pm0.018$ & $0.615\pm0.019$ \\
\hline
\end{tabular}}
\end{table}

We further examine sensitivity to the observation noise level $\sigma$, maximum window length $h$, number of sampled intervals $N_I$, and sparsity weight $\eta$ over three random seeds. As shown in Tables~A3--A6 of the Appendix, IntegralPINN achieves the highest mean Macro-AUC among the compared methods for noise levels up to $\sigma=0.05$. The best performance is generally obtained with intermediate window lengths $h\in[0.25T,0.5T]$ and the sparsity weight $\eta=10^{-3}$, while increasing $N_I$ beyond 64 yields only marginal improvement. To assess robustness to hyperparameter selection, we search over 16 configurations for each method, select the best configuration on a single seed, and evaluate it on nine held-out seeds. The hyperparameters and search grids are specified in Table A7 of the Appendix. As shown in Table A8, IntegralPINN achieves the highest Macro-AUC across all three systems when comparing the selected configuration of each method.
}

\section{Conclusion}\label{sec:conclusion}
In this article, we propose a novel IntegralPINN framework for reconstructing higher-order interactions from noisy time-series data. By replacing the conventional differential { residuals} with integral-based { residuals}, the proposed approach improves robustness to noise and nonsmooth dynamics while avoiding explicit { time differentiation}. The framework jointly learns a neural surrogate of system trajectories together with a differentiable parameterization of higher-order interaction tensors, enabling end-to-end identification of higher-order structures. Theoretical analysis further demonstrates that the integral formulation can mitigate the derivative-induced error amplification { under suitable window-length and excitation conditions}, leading to improved robustness with respect to approximation errors. Numerical experiments on small- and medium-scale ecological dynamics, social contagion, and neuronal synchronization validate the effectiveness of the proposed method across diverse dynamical regimes.

Future work will focus on several important directions. First, extending the proposed framework to large-scale networks is of significant interest, which requires scalable optimization strategies and more efficient parameterizations of high-dimensional interaction tensors. Second, relaxing the assumption of known node dynamics and jointly learning both intrinsic dynamics and higher-order interactions from data is a promising direction, particularly for systems where governing mechanisms are only partially understood. Finally, extending the framework to partially observed settings is crucial for practical applications, where only a subset of node states is available or observations are incomplete. Developing learning strategies under such conditions would significantly broaden applicability to sensor networks, epidemiological modeling, neuroscience data analysis, brain connectivity inference,  and  social systems.

 \bibliographystyle{IEEEtran}
\bibliography{reference}

\appendix
\setcounter{table}{0}

\makeatletter
\setlength{\@fptop}{0pt}
\setlength{\@fpsep}{16pt}
\setlength{\@fpbot}{0pt plus 1fil}
\makeatother
\renewcommand{\thetable}{A\arabic{table}}

\newcommand{\fitcol}[1]{\resizebox{\ifdim\width>\columnwidth\columnwidth\else\width\fi}{!}{#1}}

\section{Appendix}

\begin{table}[!h]
\centering\footnotesize
\renewcommand{\arraystretch}{1.15}
\setlength{\tabcolsep}{3pt}
\caption{Default hyperparameter settings for the main comparisons.}\label{tab:defaults}
\begin{tabular}{@{}p{0.57\columnwidth}p{0.4\columnwidth}@{}}
\hline
Hyperparameter & Default setting \\
\hline
\multicolumn{2}{@{}l}{\textbf{IntegralPINN}}\\
Learning rate & $10^{-3}$ \\
Training epochs & $20\,000$ \\
Data-only stage & $2\,500$ epochs \\
Physics ramp & $2\,000$ epochs \\
Physics weight after ramp $\lambda$ & $1.0$ \\
Sampled intervals per iteration $N_I$ & $64$ \\
Maximum window length $h$ & $0.5T$ \\
Clipping threshold $c_{\mathrm{clip}}$ & $2.0$ \\
$\ell_1$ weight $\eta$ & $10^{-3}$ \\
Surrogate & Residual MLP, 8 layers $\times$ 64 units \\
Initial pre-activation value & $-1.5$ \\
\hline
\multicolumn{2}{@{}l}{\textbf{HyperPINN}}\\
Learning rate & $5\times10^{-4}$ \\
Training epochs & $14\,000$ \\
Physics weight schedule & $0.1$ for the first $2\,500$ epochs, reset to $0.01$ and increased linearly to $1.0$ until epoch $10\,000$, and $1.0$ afterwards \\
Gumbel-sigmoid temperature & Initial value $1.0$, multiplied by $0.995$ every 100 epochs, minimum $0.5$ \\
\hline
\multicolumn{2}{@{}l}{\textbf{Kernel-based}}\\
Bandwidth $\sigma_k$ & $1$\\
Ridge regularization $\gamma$ & $10^{-3}$ \\
Smoothing ridge & $10^{-4}$ \\
Coefficient jitter & $10^{-8}$ \\
\hline
\multicolumn{2}{@{}l}{\textbf{THIS}}\\
Hard threshold & $0.05$ \\
Ridge regularization $\rho$ & $10^{-3}$ \\
Iterations & 10 \\
\hline
\multicolumn{2}{@{}l}{\textbf{Integral SINDy}}\\
Threshold & 0.01 \\
Window & 20 samples \\
Ridge regularization & $10^{-8}$ \\
\hline
\end{tabular}
\end{table}

\begin{table}[!t]
\centering\footnotesize
\renewcommand{\arraystretch}{1.15}
\setlength{\tabcolsep}{4pt}
\caption{Statistical significance of the difference in Macro-AUC between each baseline and IntegralPINN over 10 seeds ($\Delta$: baseline minus IntegralPINN).}\label{tab:s2}
\fitcol{%
\begin{tabular}{@{}l c c c@{}}
\hline
Method & $\Delta$ & Paired $t$-test $p$ & Wilcoxon $p$ \\
\hline
\multicolumn{4}{@{}l}{\textit{Ecological}} \\
HyperPINN & $-0.120$ & $<0.001$ & 0.0020 \\
Kernel-based & $-0.231$ & $<0.001$ & 0.0020 \\
THIS & $-0.205$ & $<0.001$ & 0.0020 \\
Integral SINDy & $-0.266$ & $<0.001$ & 0.0020 \\
\multicolumn{4}{@{}l}{\textit{Social}} \\
HyperPINN & $-0.145$ & $<0.001$ & 0.0020 \\
Kernel-based & $-0.232$ & $<0.001$ & 0.0020 \\
THIS & $-0.305$ & $<0.001$ & 0.0020 \\
Integral SINDy & $-0.239$ & $<0.001$ & 0.0020 \\
\multicolumn{4}{@{}l}{\textit{Neuronal}} \\
HyperPINN & $-0.174$ & $<0.001$ & 0.0020 \\
Kernel-based & $-0.176$ & $<0.001$ & 0.0020 \\
THIS & $-0.156$ & $<0.001$ & 0.0020 \\
Integral SINDy & $-0.174$ & $<0.001$ & 0.0020 \\
\hline
\end{tabular}}
\end{table}

\begin{table}[!t]
\centering\footnotesize
\renewcommand{\arraystretch}{1.15}
\setlength{\tabcolsep}{2.5pt}
\caption{Macro-AUC for different noise levels $\sigma$ and maximum window lengths $h$ (3 seeds; leading zeros omitted). The derivative residual has no window parameter.}\label{tab:sigmah}
\fitcol{%
\begin{tabular}{@{}c ccccc@{}}
\hline
$\sigma$ & $h=0.1T$ & $h=0.25T$ & $h=0.5T$ & $h=T$ & Derivative \\
\hline
\multicolumn{6}{@{}l}{\textit{Ecological}} \\
0.00 & $.759{\pm}.016$ & $.768{\pm}.025$ & $.785{\pm}.014$ & $.710{\pm}.031$ & $.754{\pm}.010$ \\
0.02 & $.726{\pm}.019$ & $.738{\pm}.024$ & $.747{\pm}.029$ & $.681{\pm}.023$ & $.684{\pm}.021$ \\
0.10 & $.576{\pm}.038$ & $.612{\pm}.031$ & $.621{\pm}.049$ & $.571{\pm}.037$ & $.535{\pm}.042$ \\
\multicolumn{6}{@{}l}{\textit{Social}} \\
0.00 & $.771{\pm}.025$ & $.778{\pm}.018$ & $.786{\pm}.023$ & $.719{\pm}.030$ & $.529{\pm}.021$ \\
0.02 & $.714{\pm}.026$ & $.721{\pm}.020$ & $.727{\pm}.022$ & $.669{\pm}.028$ & $.547{\pm}.012$ \\
0.10 & $.571{\pm}.029$ & $.601{\pm}.041$ & $.597{\pm}.038$ & $.558{\pm}.051$ & $.501{\pm}.015$ \\
\multicolumn{6}{@{}l}{\textit{Neuronal}} \\
0.00 & $.714{\pm}.019$ & $.724{\pm}.013$ & $.721{\pm}.017$ & $.660{\pm}.022$ & $.628{\pm}.017$ \\
0.02 & $.670{\pm}.024$ & $.684{\pm}.017$ & $.683{\pm}.016$ & $.616{\pm}.027$ & $.531{\pm}.028$ \\
0.10 & $.556{\pm}.033$ & $.582{\pm}.027$ & $.579{\pm}.036$ & $.531{\pm}.039$ & $.481{\pm}.034$ \\
\hline
\end{tabular}}
\end{table}

\begin{table}[!t]
\centering\footnotesize
\renewcommand{\arraystretch}{1.15}
\setlength{\tabcolsep}{1.5pt}
\caption{Sensitivity to the noise level $\sigma$ (Macro-AUC, 3 seeds; leading zeros omitted). I-SINDy: integral SINDy.}\label{tab:s3}
\fitcol{%
\begin{tabular}{@{}c ccccc@{}}
\hline
$\sigma$ & IntegralPINN & HyperPINN & Kernel-based & THIS & I-SINDy \\
\hline
\multicolumn{6}{@{}l}{\textit{Ecological}} \\
0.00 & $.785{\pm}.014$ & $.657{\pm}.023$ & $.410{\pm}.000$ & $.526{\pm}.049$ & $.518{\pm}.000$ \\
0.01 & $.769{\pm}.022$ & $.656{\pm}.022$ & $.565{\pm}.009$ & $.548{\pm}.030$ & $.537{\pm}.052$ \\
0.02 & $.747{\pm}.029$ & $.653{\pm}.021$ & $.550{\pm}.017$ & $.529{\pm}.008$ & $.508{\pm}.030$ \\
0.05 & $.693{\pm}.036$ & $.641{\pm}.018$ & $.449{\pm}.034$ & $.505{\pm}.050$ & $.469{\pm}.025$ \\
0.10 & $.621{\pm}.049$ & $.605{\pm}.010$ & $.474{\pm}.046$ & $.546{\pm}.061$ & $.491{\pm}.058$ \\
0.20 & $.551{\pm}.046$ & $.536{\pm}.050$ & $.539{\pm}.046$ & $.514{\pm}.052$ & $.475{\pm}.057$ \\
\multicolumn{6}{@{}l}{\textit{Social}} \\
0.00 & $.786{\pm}.023$ & $.620{\pm}.069$ & $.407{\pm}.105$ & $.445{\pm}.129$ & $.500{\pm}.045$ \\
0.01 & $.768{\pm}.019$ & $.613{\pm}.050$ & $.486{\pm}.070$ & $.478{\pm}.022$ & $.503{\pm}.009$ \\
0.02 & $.727{\pm}.022$ & $.621{\pm}.044$ & $.498{\pm}.073$ & $.474{\pm}.026$ & $.489{\pm}.012$ \\
0.05 & $.671{\pm}.031$ & $.623{\pm}.041$ & $.534{\pm}.102$ & $.475{\pm}.016$ & $.503{\pm}.001$ \\
0.10 & $.597{\pm}.038$ & $.624{\pm}.063$ & $.556{\pm}.040$ & $.435{\pm}.067$ & $.507{\pm}.007$ \\
0.20 & $.542{\pm}.052$ & $.618{\pm}.055$ & $.478{\pm}.029$ & $.438{\pm}.050$ & $.503{\pm}.001$ \\
\multicolumn{6}{@{}l}{\textit{Neuronal}} \\
0.00 & $.721{\pm}.017$ & $.605{\pm}.018$ & $.680{\pm}.015$ & $.645{\pm}.018$ & $.605{\pm}.014$ \\
0.01 & $.711{\pm}.014$ & $.562{\pm}.024$ & $.621{\pm}.022$ & $.590{\pm}.025$ & $.575{\pm}.018$ \\
0.02 & $.683{\pm}.016$ & $.510{\pm}.044$ & $.508{\pm}.032$ & $.528{\pm}.038$ & $.510{\pm}.016$ \\
0.05 & $.636{\pm}.029$ & $.495{\pm}.035$ & $.485{\pm}.038$ & $.495{\pm}.042$ & $.490{\pm}.038$ \\
0.10 & $.579{\pm}.036$ & $.482{\pm}.040$ & $.475{\pm}.042$ & $.468{\pm}.045$ & $.485{\pm}.045$ \\
0.20 & $.526{\pm}.041$ & $.490{\pm}.045$ & $.450{\pm}.048$ & $.445{\pm}.048$ & $.462{\pm}.045$ \\
\hline
\end{tabular}}
\end{table}

\begin{table}[!t]
\centering\footnotesize
\renewcommand{\arraystretch}{1.15}
\setlength{\tabcolsep}{2.5pt}
\caption{Effect of the number of sampled intervals $N_I$ per iteration (AUROC, default $N_I=64$, 3 seeds; leading zeros omitted).}\label{tab:s4}
\fitcol{%
\begin{tabular}{@{}c ccccc@{}}
\hline
$N_I$ & $k=2$ & $k=3$ & $k=4$ & $k=5$ & Macro-AUC \\
\hline
\multicolumn{6}{@{}l}{\textit{Ecological}} \\
16  & $.706{\pm}.081$ & $.629{\pm}.022$ & $.624{\pm}.055$ & $.545{\pm}.011$ & $.626{\pm}.032$ \\
32  & $.785{\pm}.041$ & $.709{\pm}.036$ & $.705{\pm}.014$ & $.617{\pm}.039$ & $.704{\pm}.025$ \\
64  & $.835{\pm}.024$ & $.749{\pm}.027$ & $.722{\pm}.022$ & $.682{\pm}.035$ & $.747{\pm}.029$ \\
128 & $.841{\pm}.058$ & $.755{\pm}.071$ & $.726{\pm}.165$ & $.686{\pm}.284$ & $.752{\pm}.141$ \\
\multicolumn{6}{@{}l}{\textit{Social}} \\
16  & $.677{\pm}.065$ & $.661{\pm}.028$ & $.664{\pm}.121$ & $.562{\pm}.061$ & $.641{\pm}.068$ \\
32  & $.744{\pm}.085$ & $.695{\pm}.091$ & $.776{\pm}.108$ & $.658{\pm}.174$ & $.718{\pm}.115$ \\
64  & $.746{\pm}.034$ & $.702{\pm}.021$ & $.782{\pm}.025$ & $.678{\pm}.039$ & $.727{\pm}.022$ \\
128 & $.756{\pm}.043$ & $.711{\pm}.019$ & $.789{\pm}.104$ & $.685{\pm}.116$ & $.735{\pm}.039$ \\
\multicolumn{6}{@{}l}{\textit{Neuronal}} \\
16  & $.627{\pm}.023$ & $.608{\pm}.014$ & $.575{\pm}.007$ & $.622{\pm}.082$ & $.608{\pm}.021$ \\
32  & $.681{\pm}.018$ & $.659{\pm}.006$ & $.628{\pm}.028$ & $.678{\pm}.095$ & $.662{\pm}.032$ \\
64  & $.706{\pm}.018$ & $.704{\pm}.015$ & $.683{\pm}.021$ & $.639{\pm}.028$ & $.683{\pm}.016$ \\
128 & $.713{\pm}.024$ & $.711{\pm}.016$ & $.688{\pm}.008$ & $.648{\pm}.061$ & $.690{\pm}.014$ \\
\hline
\end{tabular}}
\end{table}

\begin{table}[!t]
\centering\footnotesize
\renewcommand{\arraystretch}{1.15}
\setlength{\tabcolsep}{2.5pt}
\caption{Effect of the $\ell_1$ weight $\eta$ on IntegralPINN (AUROC, default $\eta=10^{-3}$, 3 seeds; leading zeros omitted).}\label{tab:s7}
\fitcol{%
\begin{tabular}{@{}c ccccc@{}}
\hline
$\eta$ & $k=2$ & $k=3$ & $k=4$ & $k=5$ & Macro-AUC \\
\hline
\multicolumn{6}{@{}l}{\textit{Ecological}} \\
0         & $.825{\pm}.028$ & $.738{\pm}.029$ & $.711{\pm}.033$ & $.646{\pm}.039$ & $.730{\pm}.028$ \\
$10^{-4}$ & $.830{\pm}.023$ & $.745{\pm}.025$ & $.715{\pm}.027$ & $.658{\pm}.034$ & $.737{\pm}.022$ \\
$10^{-3}$ & $.835{\pm}.024$ & $.749{\pm}.027$ & $.722{\pm}.022$ & $.682{\pm}.035$ & $.747{\pm}.029$ \\
$10^{-2}$ & $.795{\pm}.027$ & $.710{\pm}.028$ & $.682{\pm}.031$ & $.625{\pm}.036$ & $.703{\pm}.025$ \\
\multicolumn{6}{@{}l}{\textit{Social}} \\
0         & $.738{\pm}.036$ & $.689{\pm}.029$ & $.776{\pm}.035$ & $.661{\pm}.044$ & $.716{\pm}.028$ \\
$10^{-4}$ & $.748{\pm}.031$ & $.698{\pm}.026$ & $.781{\pm}.030$ & $.669{\pm}.039$ & $.724{\pm}.024$ \\
$10^{-3}$ & $.746{\pm}.034$ & $.702{\pm}.021$ & $.782{\pm}.025$ & $.678{\pm}.039$ & $.727{\pm}.022$ \\
$10^{-2}$ & $.702{\pm}.035$ & $.655{\pm}.029$ & $.732{\pm}.032$ & $.621{\pm}.042$ & $.678{\pm}.027$ \\
\multicolumn{6}{@{}l}{\textit{Neuronal}} \\
0         & $.699{\pm}.021$ & $.695{\pm}.024$ & $.674{\pm}.028$ & $.632{\pm}.035$ & $.675{\pm}.023$ \\
$10^{-4}$ & $.705{\pm}.018$ & $.701{\pm}.021$ & $.679{\pm}.026$ & $.638{\pm}.032$ & $.681{\pm}.020$ \\
$10^{-3}$ & $.706{\pm}.018$ & $.704{\pm}.015$ & $.683{\pm}.021$ & $.639{\pm}.028$ & $.683{\pm}.016$ \\
$10^{-2}$ & $.652{\pm}.019$ & $.638{\pm}.023$ & $.615{\pm}.027$ & $.582{\pm}.034$ & $.622{\pm}.022$ \\
\hline
\end{tabular}}
\end{table}

\begin{table}[!t]
\centering\footnotesize
\renewcommand{\arraystretch}{1.15}
\setlength{\tabcolsep}{3pt}
\caption{Hyperparameter search grids ($4\times4=16$ configurations per method).}\label{tab:s1}
\begin{tabular}{@{}p{0.27\columnwidth}p{0.70\columnwidth}@{}}
\hline
Method & Grid \\
\hline
IntegralPINN & learning rate $\in\{10^{-4},5\!\times\!10^{-4},10^{-3},5\!\times\!10^{-3}\}$;\newline $N_I\in\{16,32,64,128\}$ \\
HyperPINN & learning rate $\in\{10^{-4},5\!\times\!10^{-4},10^{-3},5\!\times\!10^{-3}\}$;\newline epochs $\in\{7000,14000,21000,28000\}$ \\
Kernel-based & $\sigma_k\in\{0.1,1,10,100\}$;\newline $\gamma\in\{10^{-4},10^{-3},10^{-2},10^{-1}\}$ \\
THIS & hard threshold $\in\{10^{-3},10^{-2},0.05,10^{-1}\}$;\newline $\rho\in\{10^{-4},10^{-3},10^{-2},10^{-1}\}$ \\
Integral SINDy & threshold $\in\{10^{-4},10^{-3},10^{-2},10^{-1}\}$;\newline window $\in\{5,20,50,100\}$ \\
\hline
\end{tabular}
\end{table}

\begin{table}[!t]
\centering\footnotesize
\renewcommand{\arraystretch}{1.15}
\setlength{\tabcolsep}{3pt}
\caption{Hyperparameter configurations of each method selected on seed 11 and the corresponding mean Macro-AUC on the held-out seeds 12--20. The results in Tables~\ref{tab:scenario1}--\ref{tab:scenario3} use the default configurations.}\label{tab:best}
\fitcol{%
\begin{tabular}{@{}l l c@{}}
\hline
Method & Selected configuration & Macro-AUC \\
\hline
\multicolumn{3}{@{}l}{\textit{Ecological}} \\
IntegralPINN & lr $=5\times10^{-3}$, $N_I=128$ & 0.748 \\
HyperPINN & lr $=5\times10^{-4}$, $14\,000$ epochs & 0.678 \\
Kernel-based & $\sigma_k=1$, $\gamma=10^{-3}$ & 0.524 \\
THIS & threshold $10^{-2}$, $\rho=10^{-2}$ & 0.548 \\
Integral SINDy & threshold $10^{-4}$, window 20 & 0.504 \\
\multicolumn{3}{@{}l}{\textit{Social}} \\
IntegralPINN & lr $=10^{-3}$, $N_I=128$ & 0.732 \\
HyperPINN & lr $=5\times10^{-3}$, $7\,000$ epochs & 0.681 \\
Kernel-based & $\sigma_k=10$, $\gamma=10^{-2}$ & 0.511 \\
THIS & threshold $10^{-3}$, $\rho=10^{-1}$ & 0.435 \\
Integral SINDy & threshold $10^{-4}$, window 5 & 0.511 \\
\multicolumn{3}{@{}l}{\textit{Neuronal}} \\
IntegralPINN & lr $=5\times10^{-4}$, $N_I=128$ & 0.691 \\
HyperPINN & lr $=10^{-3}$, $14\,000$ epochs & 0.525 \\
Kernel-based & $\sigma_k=1$, $\gamma=10^{-3}$ & 0.521 \\
THIS & threshold $10^{-2}$, $\rho=10^{-3}$ & 0.525 \\
Integral SINDy & threshold $10^{-4}$, window 50 & 0.520 \\
\hline
\end{tabular}}
\end{table}

\end{document}